\documentclass[11pt,a4paper]{article}
\usepackage[hyperref]{acl}
\usepackage{times}
\usepackage{latexsym}
\usepackage{amsmath,amssymb,amsthm,amsfonts}
\usepackage{graphicx}
\usepackage{booktabs}
\usepackage{subcaption}
\usepackage{url}
\usepackage{wrapfig}
\usepackage{microtype}
\usepackage{inconsolata}
\usepackage{svg}

\usepackage[most]{tcolorbox}
\newtcbox{\mybox}[1][red]{on line,
    colback=#1, colframe=#1, boxsep=-5pt, boxrule=0pt, size=small, arc=1mm}
\usepackage{fancybox}

\newif\ifcomment
\commenttrue

\ifcomment
\newcommand{\rf}[1]{\textcolor{red}{RF: #1}}
\newcommand{\ip}[1]{\textcolor{blue}{IP: #1}}
\newcommand{\km}[1]{\textcolor{orange}{KM: #1}}
\newcommand{\ec}[1]{\textcolor{cyan}{EC: #1}}
\else
\newcommand{\rf}[1]{}
\newcommand{\ip}[1]{}
\newcommand{\km}[1]{}
\newcommand{\ec}[1]{}
\fi

\usepackage{todonotes}
\makeatletter
\newcommand*\iftodonotes{\if@todonotes@disabled\expandafter\@secondoftwo\else\expandafter\@firstoftwo\fi}  
\makeatother

\title{When classifying grammatical role, BERT doesn't care about word order\dots except when it matters}

\author{Isabel Papadimitriou\\
  Stanford University\\
  \texttt{isabelvp@stanford.edu} \\\And
  Richard Futrell\\
  University of California, Irvine\\
  \texttt{rfutrell@uci.edu} \\\AND
  Kyle Mahowald\\
  The University of Texas at Austin\\
  \texttt{mahowald@utexas.edu}}

\date{}

\begin{document}
\maketitle


\begin{abstract}
    Because meaning can often be inferred from lexical semantics alone, word order is often a redundant cue in natural language. For example, the words \textit{chopped}, \textit{chef}, and \textit{onion} are more likely used to convey ``The chef chopped the onion," not ``The onion chopped the chef." Recent work has shown large language models to be surprisingly word order invariant, but crucially has largely considered natural \textit{prototypical} inputs, where compositional meaning mostly matches lexical expectations. To overcome this confound, we probe grammatical role representation in English BERT and GPT-2, on instances where lexical expectations are not sufficient, and word order knowledge is necessary for correct classification. Such \textit{non-prototypical} instances are naturally occurring English sentences with inanimate subjects or animate objects, or sentences where we systematically swap the arguments to make sentences like ``The onion chopped the chef".  We find that, while early layer embeddings are largely lexical, word order is in fact crucial in defining the later-layer representations of words in semantically non-prototypical positions. Our experiments isolate the effect of word order on the contextualization process, and highlight how models use context in the uncommon, but critical, instances where it matters. 
\end{abstract}

\section{Introduction and Prior Work}

Large language models create contextual embeddings of the words in their input, starting with a static embedding of each token and progressively adding more contextual information in each layer \citep{devlin-etal-2019-bert,brown2020language,manning2020emergent}. 
While these contextual embedding models are often praised for capturing rich grammatical structure, a spate of recent work has shown that they are surprisingly invariant to scrambling word order \citep{sinha2021masked,hessel2021effective,pham2020out,gupta2021bert,o2021context} and that grammatical knowledge like part of speech, often attributed to contextual embeddings, is actually also captured by fixed embeddings \citep{pimentel2020information}. These results point to a puzzle: how can syntactic contextual information be important for language understanding when the words themselves, not their order, are what matter?

\begin{figure}
    \centering
    \includegraphics[width=\columnwidth]{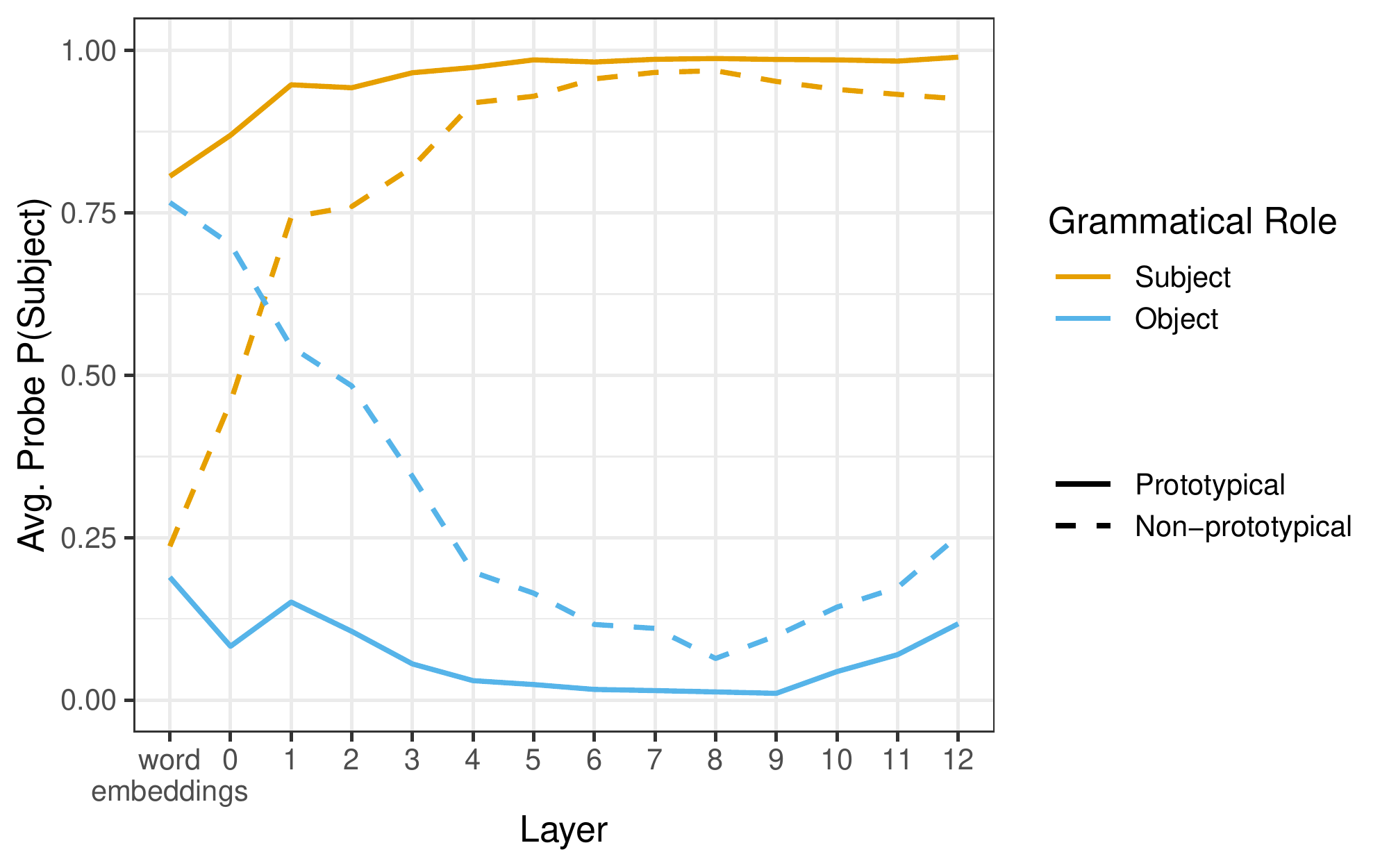}
    \caption{Probabilities of probes trained to differentiate subjects from objects in BERT embeddings. We separate our evaluation examples by prototypicality: whether the ground truth grammatical role is what we would expect given the word out of context. The majority of natural examples are prototypical (solid lines), and so if we average all cases we cannot see that grammatical information is gradually acquired in the first half of the network for cases where lexical information is non-prototypical. The equivalent figures for GPT-2  are in Appendix \ref{gpt2_figs}.}
    \label{fig:prototypical}
\end{figure}

We argue that this apparent paradox arises because of the redundant structure of language itself. Lexical distributional information alone inherently captures a great deal of meaning \citep{erk2012vector,mitchell2010composition,tal2022redundancy}, and typically both humans and machines can reconstruct meanings of sentences under local scrambling of words \citep{mollica2020composition,clouatre2021demystifying}. In this paper, we study model behaviour in cases where word order is informative and \textit{is not} redundant with lexical information. 

We focus on the feature of \textbf{grammatical role} (whether a noun is the subject or the object of a clause). Most natural clauses are \textbf{prototypical}: in a sentence like ``the chef chopped the onion'', the grammatical roles of \emph{chef} and \emph{onion} are clear to humans from the words alone, without word order or context \citep[see][for experiments in English and Russian in which human participants successfully guessed which of two nouns was the subject and which was the object of a simple transitive clause, in the absence of word order and contextual information]{mahowald2022grammatical}.
This means syntactic word order is often redundant with lexical semantics.
Whether hand-constructed or corpus-based, most studies probing contextual representations have used prototypical sentences as input, where syntactic word order may not have much information to contribute to core meaning beyond the words themselves. 

Yet human language can use syntax to deviate from the expectations generated by lexical items: we can also understand the absurd meaning of a rare \textbf{non-prototypical} sentence like ``The onion chopped the chef'' \citep{garrett1976syntactic,gibson2013noisy}. Is this use of syntactic word order available to pretrained models?  In this paper, we train grammatical role probes on the embedding spaces of BERT and GPT-2\footnote{Results are similar for the two models, so we visualize BERT results here, and include GPT-2 figures in Appendix \ref{gpt2_figs}.}, and evaluate them
on these rare non-prototypical examples, where the meaning of words in context is different from what we would expect from looking at the words alone.

We probe for grammatical role because it is key to the basic compositional semantic structure of a sentence \citep{dixon1979ergativity,comrie1989language,croft2001functional}. 
 While fixed lexical semantics contains information about grammatical role (animate nouns are likely to be subjects, etc), the grammatical role of a word in English is ultimately determined by syntactic word order. Probing grammatical role lets us examine the interplay between syntactic word order and lexical semantics in forming compositional meaning through model layers. 
 
For all of our experiments, we train grammatical role probes with standard data and test them on either prototypical cases or non-prototypical cases (where word order matters), to understand if grammatical embedding under normal circumstances is sensitive to word order. Our experiments reveal three key findings: 
\begin{enumerate}
    \item Lexical semantics plays a key role in organizing embedding space in early layer representations, and  non-lexical compositional features are expressed gradually in later layers, as shown by probe performance on non-prototypical sentences (Experiment 1, Figure \ref{fig:prototypical}). 
    \item Embeddings represent meaning that is imparted \textit{only} by syntactic word order, overriding lexical and distributional cues. When we control for distributional co-occurrence factors by evaluating our probes on \textbf{argument swapped sentences} (like ``The onion chopped the chef'', real sample in Appendix \ref{sec:sample}), probes can differentiate the same word in different roles  (Experiment 2, Figure \ref{fig:swapped}).
    \item Syntactic word order is significant beyond just local coherence: the compositional information of syntactic word order is lost when we test our probes on locally-shuffled sentences, that keep local lexical coherence but break acute syntactic relations (Figure \ref{fig:localshuf}).
\end{enumerate}
More generally, we highlight the importance of examining models using non-prototypical examples, both for understanding the strength of lexical influence in contextual embeddings, but also for accurately isolating syntactic processing where it is taking place.\footnote{The code to run our experiments is at \url{https://github.com/toizzy/except-when-it-matters}}

%

\section{Why non-prototypical probing?}
As opposed to more general syntactic probing tasks (e.g., dependency parsing), grammatical role is a linguistically significant yet specific task that is both syntactic \textit{and} semantic. As such, we can choose these linguistically-informed sets of non-prototypical examples where the lexical semantics does not match the compositional meaning implied by the syntax. 

Non-prototypical examples give us a unique perspective on how syntactic machinery like word order influences compositional meaning representation \textit{independently} from lexical semantics. 
Studies in probing have controlled for lexical semantics by substituting content words for nonce words \citep[``jabberwocky" sentences, as in][]{maudslay2021syntactic,goodwin2020probing} or random real words \citep[``colorless green idea" sentences, as in][]{gulordava-etal-2018-colorless}.
A tradeoff is that these methods lead to out-of-distribution sentences whose words are unlikely to ever co-occur naturally.
Rather than bleaching the effect of lexical semantics, our setup lets us examine the interplay between lexical semantics and syntactic representation in a controlled environment, isolating the effects of syntactic word order while using in-distribution examples.

Recent work on representation probing has focused on improving probing methodologies to make sure that extracted information is not spurious or not simply lexical \citep{hewitt-liang-2019-designing,belinkov2022probing,voita2020information,hewitt2021conditional,pimentel2020information}.
Our experiments are a complementary approach, where we use standard probing methods, but use linguistically-informed \textit{data selection} to address the ambiguity of what classifiers are extracting.

\section{Experiment 1: Grammatical Subjecthood Probes}

In Experiment 1, we evaluate grammatical role probes on prototypical instances, where grammatical role lines up with lexical expectations, and non-prototypical instances, where it does not.

\subsection{Methods}

We train a 2-level perceptron classifier probe with 64 hidden units to distinguish the layer embeddings of nouns that are \textit{transitive subjects} from nouns that are \textit{transitive objects}, as in \citet{papadimitriou-etal-2021-deep}. We train a separate classifier for each model layer, as well as training a classifier on the static word embedding space of the models without the position embeddings added (before layer 0). The probe classifiers are binary, taking the layer embedding of a noun and predicting whether it is a transitive subject or a transitive object. Probe training data comes from Universal Dependencies treebanks: we pass single sentences from the treebanks through the models, and use dependency annotations to label each layer embedding for whether it represents a transitive subject, a transitive object, or neither (not included in training). The training set is balanced, and consists of 864 embeddings of subject nouns, and 864 embeddings of object nouns. We train all probes for 20 epochs, for consistency. The embedding models that we use are \texttt{bert-base-uncased} and \texttt{gpt2}.
For our analysis, we call a noun a prototypical subject if the probe probability for its word embedding (pre-layer 0) is greater than $0.5$, and a prototypical object if it is less.

\subsection{Results}

Prototypical and non-prototypical arguments differ in probing behavior across layers, as demonstrated in Figure \ref{fig:prototypical}.
For prototypical instances (solid lines), syntactic information is conflated with type-level information and so probe accuracy is high starting from layer 0 (word embeddings + position embeddings), and stays consistent throughout the network. 
However, when we look at non-prototypical instances (dashed lines), we see that the embeddings from layer to layer have very different grammatical encodings, with type-level semantics dominating in the early layers and more general syntactic knowledge only becoming extractable by our probes in later layers. 

Crucially, since prototypical examples dominate in frequency in any corpus, the average probe accuracy across all examples is high for all layers, and the grammatical encoding of subjecthood, which is accurate only after the middle layers of the model, would be hidden.
Separating out non-prototypical examples illustrates how the syntax of a phrase can arise independently from type-level information through transformer layers, while also showcasing the importance of lexical semantics in forming embedding space geometry in the first half of the network.

\section{Experiment 2: Controlling for Distributional Information by Swapping Subjects and Objects}

In Experiment 1 we show that the contextualization process consists of gradual grammatical information gain for non-prototypical examples, even though this is largely obscured in the majority prototypical examples where the lexical semantics also contains accurate syntactic information. In this experiment, we ask: does this contextualized information about grammatical role stem from word order and syntax, or from distributional (bag-of-words) effects when seeing all words in the sentence? We answer this question by creating example pairs where we control for distributional information by keeping all the words the same, but swapping the positions of the subject and the object. Such pairs of the type ``The chef chopped the onion'' $\rightarrow$ ``The onion chopped the chef'' (real sample in Appendix \ref{sec:sample}) have identical distributional information. To accurately classify grammatical role in both sentences, the model we're probing would have to be attuned to the ways in which small changes in word order globally affect meaning. 
\begin{figure}
    \centering
    \includegraphics[width=\columnwidth]{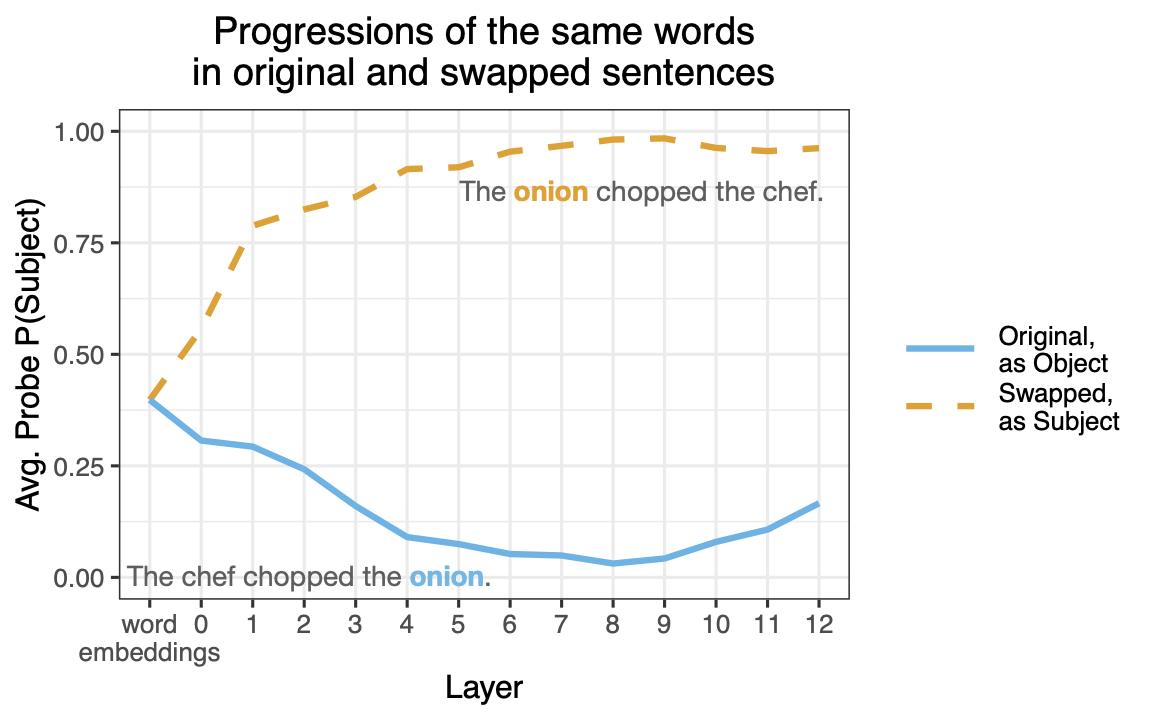} 
    \caption{Average probe probabilities for our argument-swapped test set. We visualize the probabilities for the same words when they are the object of an original treebank sentence (eg. ``The chef chopped the \textbf{onion}'', blue line) versus being the subject of that sentence after manual swapping (eg. ``The \textbf{onion} chopped the chef'', dashed red line). When probing the geometry of grammatical role, \textit{the same words in the same distributional contexts} are clearly differentiated throughout contextualization in BERT layers, due to the impact of syntactic word order.}
    \label{fig:swapped}
\end{figure}

\subsection{Methods}
\label{sec:swapmethods}
We use the same probing classifiers from Experiment 1, and evaluate on a special test set of pairs of sentences that have the subject and direct object of one clause swapped. To create the swapped sentences, we search the UD treebank for verbs that have lexical, non-pronoun direct subjects and direct objects, check that the subject and object have the same number (singular or plural), and also check that neither of them are part of a compound word or a flat dependency word that would be separated (like a full name). If a sentence contains a verb where its arguments fulfill all of these requirements, we swap the position of the subject and the object to create a second, swapped sentence, and add the sentence pair (original and swapped) to our evaluation set\footnote{We do not filter for prototypical subjects and objects in this process, since we are assessing the effect of all distributional information: a sentence like ``The onion made the chef cry'' has nouns in non-prototypical roles, but is still much more felicitous than its swapped version}. A random sample of our swapped sentences is in Appendix \ref{sec:sample}.

\subsection{Results}

When testing our probes on pairs of normal and swapped sentences, we find that our probes from Experiment 1 correctly classify both the normal and the swapped sentences with high accuracy in higher layers. 
Since we test our probes on controlled pairs that have the same distributional information, we can isolate effect of syntactic word order in influencing meaning representation. This is demonstrated in Figure \ref{fig:swapped}, where probe predictions for the same set of words in the same distributional context diverges significantly depending on whether the word is in subject or object position.
Our results indicate that, separate from distributional effects, models have learnt to represent the ways in which syntactic word order can \textit{independently} affect meaning.

\begin{figure}
    \centering
    \includegraphics[width=\columnwidth]{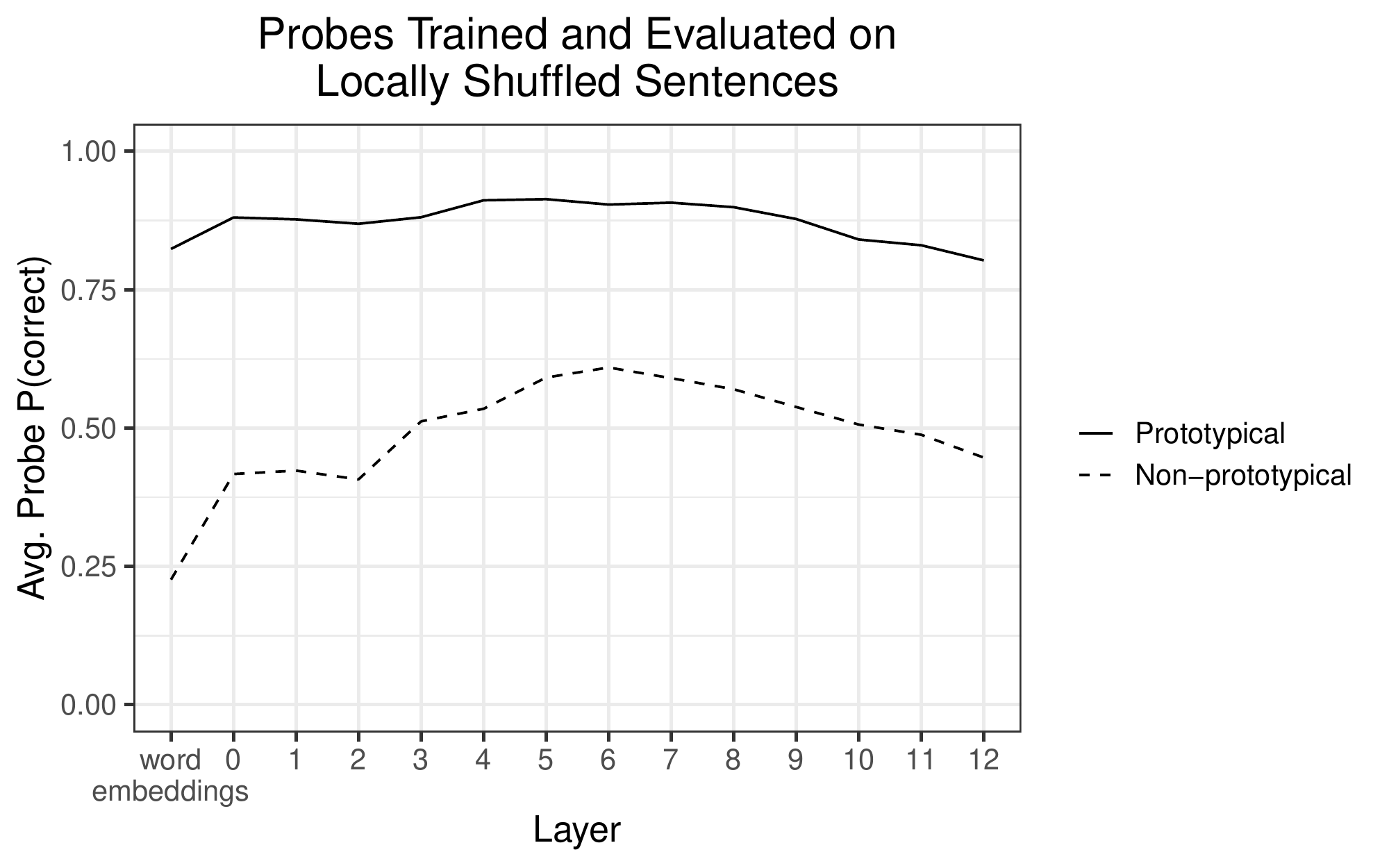}
    \caption{ Probe accuracies for sentences where the words have been locally scrambled such that no word moves more than 2 slots. Probe performance for non-prototypical sentences is close to chance, indicating that general positional information (still available after local scrambling) is not enough to recover grammatical role. However, the lexical semantics is preserved through layers in these scrambled instances as evidenced by the steady probe performance on prototypical sentences. 
    }
    \label{fig:localshuf}
\end{figure}

\subsection{Are these results just due to general position information?}

Our results in Experiment 2 indicate that syntactic word order information can affect model representations of word meaning, even when we keep lexical and distributional information constant. A question still remains: does the divergence demonstrated in Figure \ref{fig:swapped} stem from the fine-grained ways in which word order influences syntax in English, or from heuristics based on primacy (whether a word is earlier or later in a sentence)? To further investigate this, we train and test probes on sentences where word order is locally scrambled so that no word moves more than 2 slots, and so general primacy and local coherence is preserved.
As shown in Figure \ref{fig:localshuf}, probes trained on these locally shuffled sentences do not fare better than chance on non-prototypical examples. While prototypical lexical information can aid classification (solid line), general primacy information is not sufficient to overcome lexical cues and cause the word-order-dependent representation we demonstrate in Figure \ref{fig:swapped}.

\section{Discussion}

While recent work has shown that large language models come to rely largely on distributional semantic information, we consider the model's ability to \textit{overcome} these distributional cues. Research showing that models rely on lexical and distributional information is not at odds with our findings that this can be overridden. In fact, even though humans can accurately understand non-prototypical sentences, human syntactic processing is often influenced by the lexical semantics of words, as evidenced by studies on human subjects \citep{frazier1982making,rayner1983interaction,ferreira1990use} as well as by lexically-influenced syntactic processes in human languages, like differential object marking \citep{aissen2003differential}---a phenomenon whereby non-prototypical grammatical objects are marked.

More generally, while we have shown that it is tempting for a straightforward probing approach to conclude that grammatical role information is available to the lowest layers of BERT, 
separately analyzing prototypical and non-prototypical arguments makes it clear that the picture is more complicated.
At lower layers, BERT representations can \emph{typically} classify subjects and objects, but when a non-prototypical meaning is expressed, accurate classification is not available until the higher layers.

We argue that considering probing performance on these non-prototypical instances is crucial.
A key design feature of human language is the ability to talk about things that aren't there or don't exist \citep{hockett1960origin}, and it has been argued that the combinatoric power of syntax exists to allow humans to say things that are subtle, surprising, or impossible \citep{garrett1976syntactic,chomsky1957syntactic}.
Thus, considering probing accuracy on the \textit{average} task may be misleading.
Insofar as being able to understand non-prototypical meanings is a hallmark of human language and insofar as these meanings may differ in systematic ways from prototypical meanings, considering such cases is crucial for understanding how language models represent language.

\section{Acknowledgments}

This work was supported by National Science Foundation Grants No. 2104995 to KM, No. 1947307 to RF, and a Graduate Research Fellowship to IP. We thank Dan Jurafsky and Adina Williams for helpful discussions, and Kaitlyn Zhou, Dallas Card, and J. Adolfo Hermosillo for comments on drafts.

\bibliography{everything,references,anthology,acl2020}
\bibliographystyle{acl_natbib}

\appendix

\section{Figures for GPT-2 Experiments}
\label{gpt2_figs}

\begin{figure}
    \centering
    \includegraphics[width=\columnwidth]{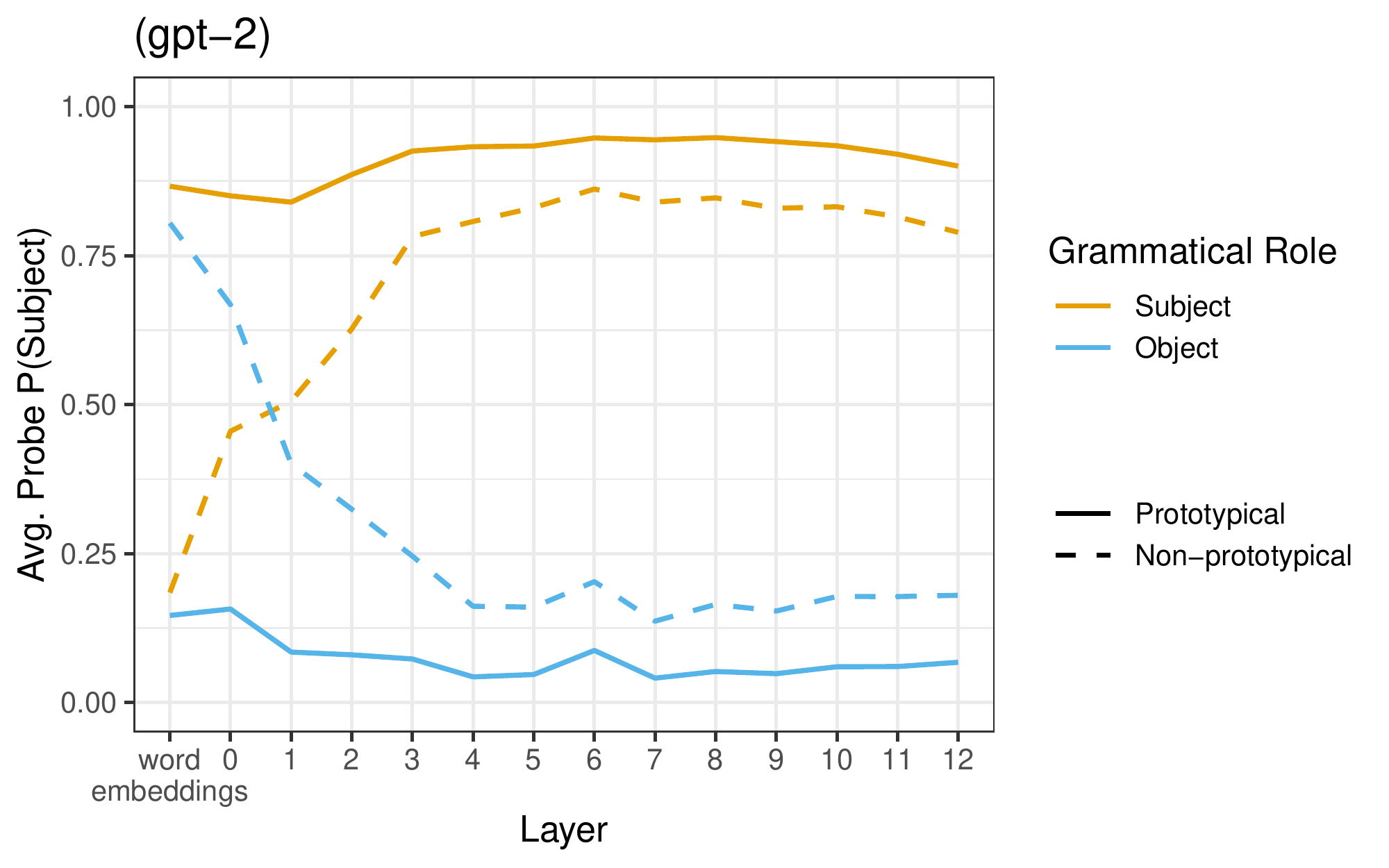}
    \caption{Equivalent to Figure \ref{fig:prototypical} from the main paper, on GPT-2 embeddings}.
    \label{fig:gpt2_fig1}

    \centering
    \includegraphics[width=\columnwidth]{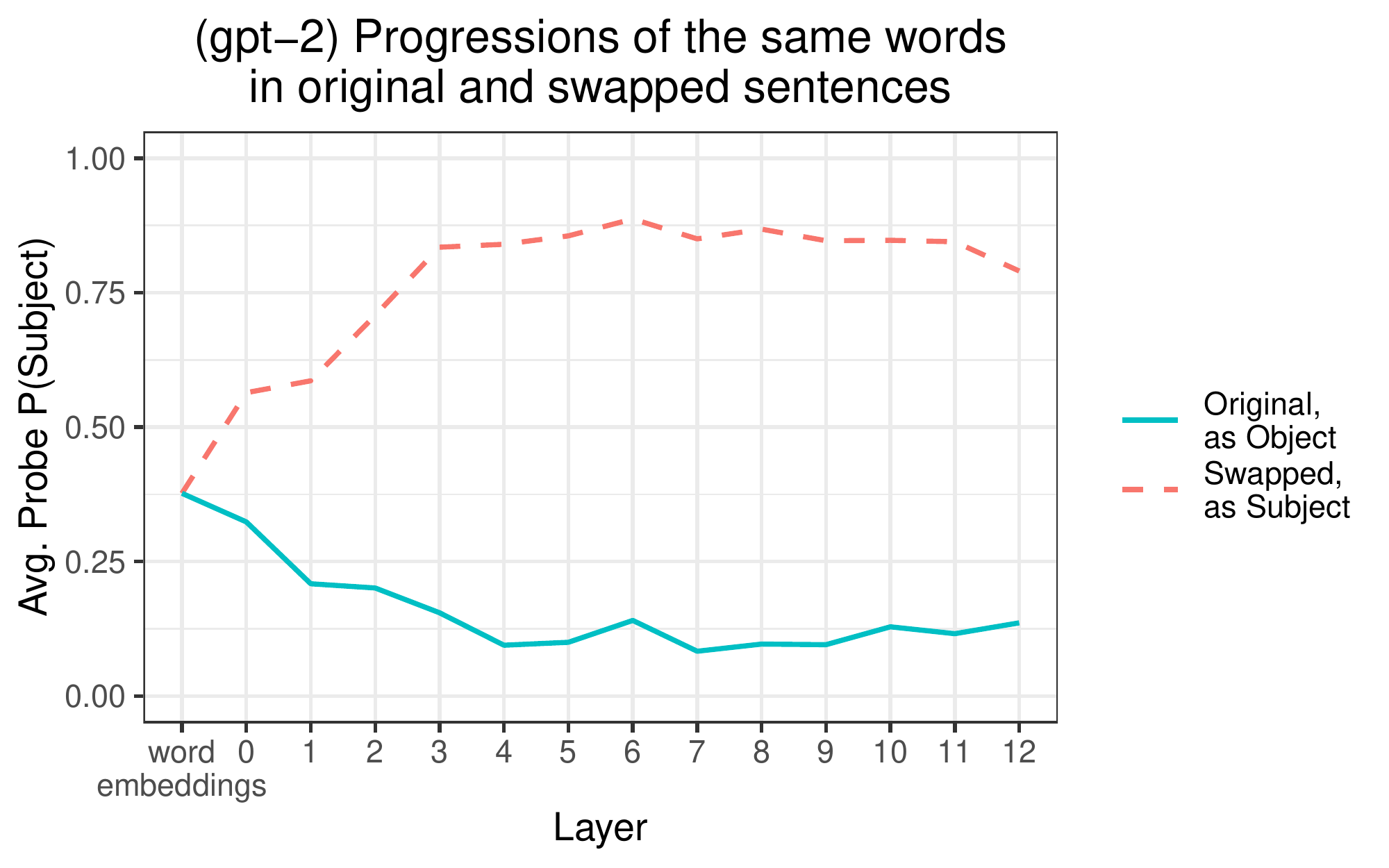}
    \caption{Equivalent to Figure \ref{fig:swapped} from the main paper, on GPT-2 embeddings. Grammatical representation in GPT-2 embedding also diverges for the same words in the same distributional contexts.}.
    \label{fig:gpt2_fig2}

    \centering
    \includegraphics[width=\columnwidth]{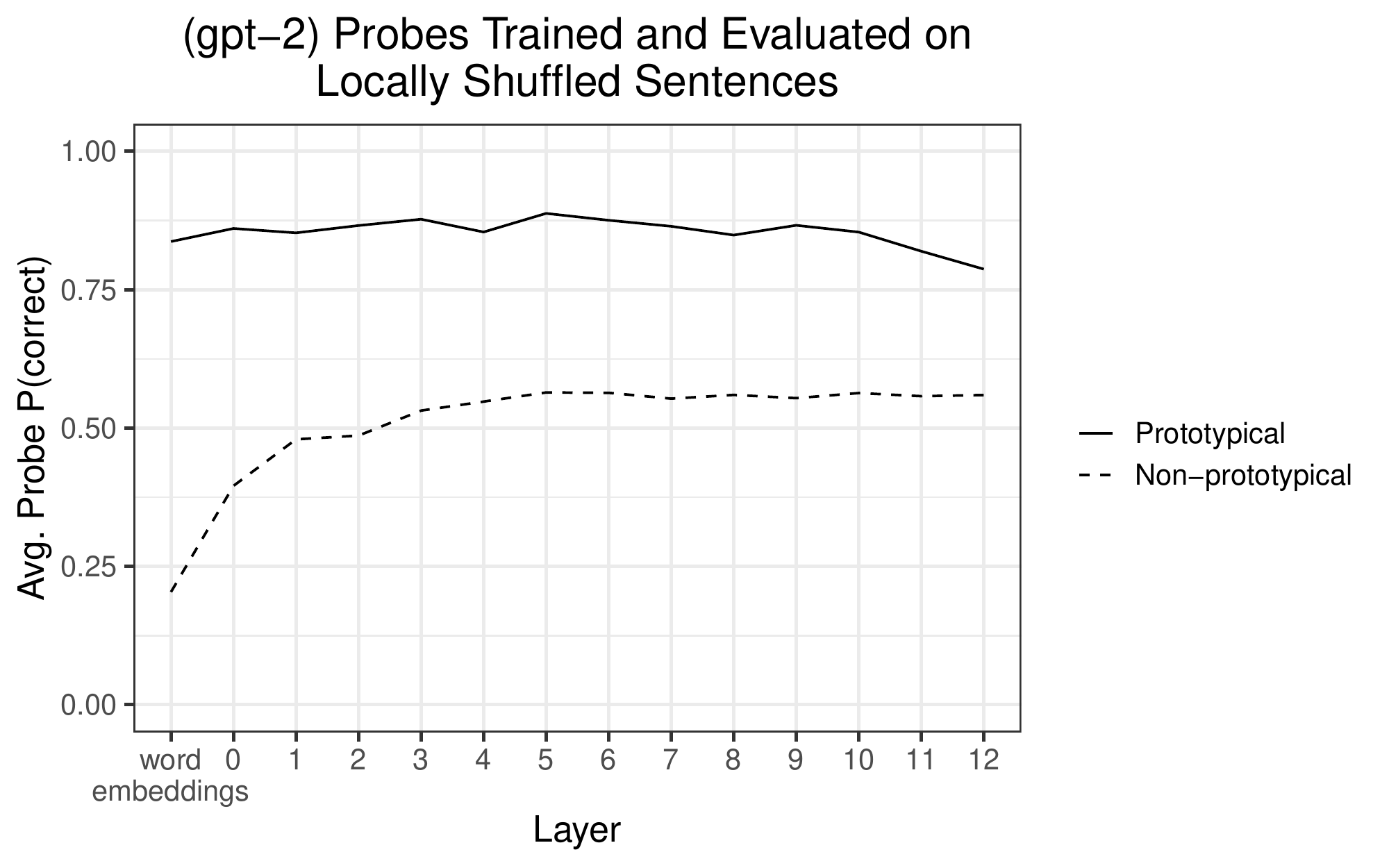}
    \caption{Equivalent to Figure \ref{fig:localshuf} from the main paper, on GPT-2 embeddings. As shown by the dashed line being close to chance, grammatical role information is not extractable from locally shuffled sentences in the non-prototypical cases where lexical semantics do not help}.
    \label{fig:gpt2_fig3}
\end{figure}

We ran our experiments on both BERT and GPT-2 embeddings, and both models had similar behaviors that we discuss in the paper. For clarity, figures in the paper only visualize the BERT results, and we're including the GPT-2 versions of those same figures for comparison. Figure \ref{fig:gpt2_fig1} shows the GPT-2 results of Figure \ref{fig:prototypical}, Figure \ref{fig:gpt2_fig2} shows the GPT-2 results of Figure \ref{fig:swapped}, and Figure \ref{fig:gpt2_fig3} shows the GPT-2 result of Figure \ref{fig:localshuf}.

\section{Sample of argument-swapped sentences}
\label{sec:sample}
A random sample (not cherry-picked) of our argument-swapped evaluation set, where the subject and the object of clauses are automatically swapped. The original subject is in \textbf{bold} and the original object is in \textit{\textbf{bold and italics}}. The process for creating these sentences is detailed in Section \ref{sec:swapmethods}

On Thursday, with 110 days until the start of the 2014 Winter Paralympics in Sochi, Russia, \textit{\textbf{Professor}} interviewed Assistant \textbf{Wikinews} in Educational Leadership, Sport Studies and Educational / Counseling Psychology at Washington State University Simon Ličen about attitudes in United States towards the Paralympics.

This \textbf{\textit{approach}} shows a more realistic \textbf{video} to playing Quidditch.

Second, aggregate \textit{\textbf{view}} provides only a high-level \textbf{information} of a field, which can make it difficult to investigate causality [23].

A \textit{\textbf{hand}} raises her \textbf{girl}.

\textbf{\textit{area}} of the Mississippi River and the destruction of wetlands at its mouth have left the \textbf{Alteration} around New Orleans abnormally vulnerable to the forces of nature.

It was known that a moving \textit{\textbf{energy}} exchanges its kinetic \textbf{body} for potential energy when it gains height.

Thus, when ACPeds issued a statement condemning gender reassignment surgery in 2016 [21], many \textit{\textbf{beliefs}} mistook the organization ’s political \textbf{people} for the consensus view among United States pediatricians — although the peak body for pediatric workers, the American Academy of Pediatrics, has a much more positive view of gender dysphoria [22].

His \textit{\textbf{painting}} perfectly combines \textbf{art} and Chinese calligraphy.

When the \textit{\textbf{inches}} become a few \textbf{plants} tall and their leaves mature, it 's time to transplant them to a larger container.

Since the television series’ inception, \textit{\textbf{reviews}} at The AV Club have written two critical \textbf{writers} for each episode:

\end{document}